\pdfoutput=1

\documentclass[11pt]{article}

\usepackage{acl}

\usepackage{tgtermes}
\usepackage{latexsym}
\usepackage{bm} 
\usepackage[T1]{fontenc}

\usepackage[utf8]{inputenc}

\usepackage{microtype}

\usepackage{paralist} 
\usepackage{amsmath}  
\usepackage{amsthm}
\usepackage{amsfonts}
\usepackage{url}      
\usepackage[vlined]{algorithm2e} 
\usepackage{setspace} 
\usepackage{listings} 
\usepackage{floatrow} 
\usepackage{graphicx}
\usepackage{boldline}
\usepackage{adjustbox}

\title{Think Twice: Measuring the~Efficiency
of~Eliminating~Prediction~Shortcuts~of~Question~Answering~Models}

\author{Lukáš Mikula$^{\clubsuit *}$ \and Michal Štefánik$^{\clubsuit *}$ \and Marek Petrovič$^\clubsuit$ \and Petr Sojka$^\clubsuit$\\
        \\$^\clubsuit$Faculty of Informatics,\\\vspace{10pt}
        Masaryk University, Czech Republic}
        
\begin{document}
\maketitle
\def\thefootnote{*}\footnotetext{First two authors contributed equally}\def\thefootnote{\arabic{footnote}}

\begin{abstract}
While the Large Language Models (LLMs) dominate a majority of language understanding tasks, previous work shows that some of these results are supported by modelling spurious correlations of training datasets. Authors commonly assess model robustness by evaluating their models on out-of-distribution (OOD) datasets of the same task, but these datasets might \textit{share} the bias of the training dataset.

We propose a simple method for measuring a scale of models' reliance on any identified spurious feature and assess the robustness towards a large set of known and newly found prediction biases for various pre-trained models and debiasing methods in Question Answering (QA). 
We find that while existing debiasing methods \textit{can} mitigate reliance on a chosen spurious feature, the OOD performance gains of these methods can \textit{not} be explained by mitigated reliance on biased features, suggesting that biases are \textit{shared} among different QA datasets. Finally, we evidence this to be the case by measuring that performance of models trained on different QA datasets relies \textit{comparably} on the \textit{same} bias features. We hope these results will motivate future work to refine the reports of LMs' robustness to a level of adversarial samples addressing specific spurious features.
\end{abstract}

\section{Introduction}
\label{chap:intro}

\begin{figure}[th]
  \centering
    \!\!\includegraphics[width=1.03\textwidth,
    keepaspectratio]{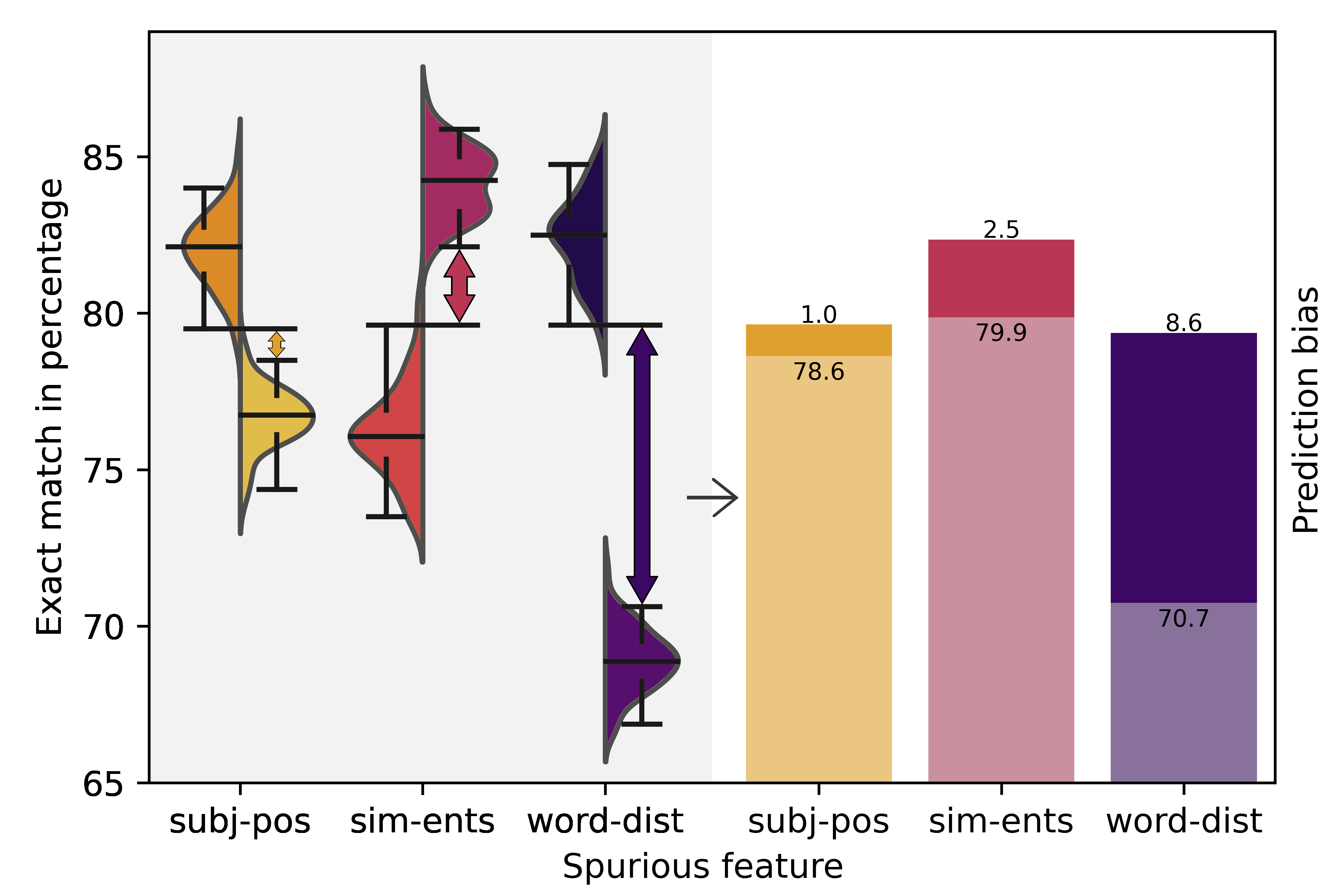}
  \vspace*{-1\baselineskip}
  \caption{We quantify model reliance on a spurious feature using bootstrapped evaluation on segments of data separated by exploiting chosen bias (left) and subsequently, by measuring the difference in model's performance over these two groups (right), that we refer to as \textit{Prediction bias} (§\ref{sec:measusing}). \vspace*{-1\baselineskip}}
  \label{fig:visual_abstract}
\end{figure}

Unsupervised pre-training and vast parametrization \cite{devlin-etal-2019-bert,Radford2018gpt} enable Large Language Models (LLMs) to reach close-to-human accuracy on complex downstream tasks such as Natural Language Inference, Sentiment Analysis, or Question Answering. 
However, previous work shows that these outstanding results can partially be attributed to models' reliance on non-representative patterns in training data shared with the test set, such as the high lexical intersection of the entailed hypothesis to premise~\cite{multitask_eliminates_biases} in Natural Language Inference (NLI) or the intersection of the question and answer vocabulary~\cite{qa:shinoda-etal-2021-question} in extractive Question Answering (QA). 

A primary motivation for mitigating models' reliance on such features is to enhance their \textit{robustness} in practice, avoiding fragility to systematic errors when responding the open-ended user requests. 
Models' robustness is commonly assessed by measuring prediction quality on samples from other out-of-distribution (OOD) datasets~\cite{clark2019don,karimi-mahabadi-etal-2020-end,Utama2020TowardsDN,xiong2021uncertainty}. 
However, the OOD datasets might \textit{share} training biases introduced by shared features, 
such as data collection methodology, or human annotators' background \cite{Mehrabi2021ASO}. In such cases, conversely, a model reliant on biased correlations can reach \textit{higher} OOD score despite being more fragile to the adversarial inputs exploiting the biased correlation.

With this motivation, we propose a framework to evaluate models' reliance on a biased feature in prediction by \textit{splitting} evaluation data to two groups based on a biased feature and \textit{comparing} the prediction quality on these two groups (Fig.~\ref{fig:visual_abstract}).
This way, we assess a reliance on bias of diverse QA models for several previously and newly identified bias features identified in this work.
Finally, we assess the efficiency of the state-of-the-art debiasing methods in mitigating reliance on spurious features over a resampling baseline and compare the findings to the commonly assessed OOD performance. 

We find that avoiding reliance on spurious features does \textit{not imply} improvements in OOD performance; in many cases, debiasing methods mitigate the model's prediction bias, but the OOD performance drops, while counterintuitively, a magnification of bias reliance can also bring large OOD gains. Aiming to explain this, we directly evaluate the prediction bias of models trained on different datasets and confirm that even models trained on OOD datasets often rely on the \textit{same} spurious correlations \textit{comparably} to the ID models. This finding motivates the presented assessment of model robustness towards known biases, in addition to OOD performance.

This paper is structured as follows. Section~\ref{sec:background} overviews data biases observed in NLP datasets, recent debiasing methods, and the previous methods related to measuring inclination to spurious correlations. Section~\ref{sec:measusing} presents our method for measuring the significance of specific biases. We follow in Section~\ref{sec:experiments} with details on our evaluation setup, including the tested debiasing methods, addressed bias features, and the design of a set of heuristics that can exploit them. Subsequently, in Section~\ref{sec:results}, we measure and report models' robustness to biases and OOD datasets before and after applying the selected debiasing methods and wrap up our observations in Sections~\ref{sec:discussion} and~\ref{sec:conclusion}.

\vspace{-.3\baselineskip}
\paragraph{Problem definition}
\label{sec:definitions}
Given a set of inputs $X=x_{1..i}$ with corresponding labels $Y=y_{1..i}$ from a dataset $\mathcal{D}_\textit{ID}$, a model $M$ learns a \textit{task} $\mathcal{T}$ by identifying \textit{features} $\mathcal{F}_{1..n}$ that map each $x_j$ to a corresponding $y_j$, assuming that the learned features must be \textit{consistent} with $\mathcal{D}_\textit{ID}$. 
Since the learned $\mathcal{F}_{1..n}$ are distributed in $M$ and can not be directly evaluated, we assess whether the learned features are \textit{robust} for the task $\mathcal{T}$ by evaluating $M$ on samples $X_\textit{OOD}$ of the same task, but drawn from $\mathcal{D}_\textit{OOD} \not\approx \mathcal{D}_\textit{ID}$; 
we assume that if $\mathcal{F}_{1..n} \in M$ are robust, the model will also perform well on $X_\textit{OOD}$. 
However, the consistency of the learned $\mathcal{F}_k$ with both $X_\textit{ID}$ and $X_\textit{OOD}$ is merely a \textit{necessary} and not a \textit{sufficient} condition for $\mathcal{F}_k$ to be robust; 
If there exists a pair $(x, y)$ such that the pair is a \textit{valid} sample of the task $\mathcal{T}$, but is not consistent with $\mathcal{F}_k$, we denote $\mathcal{F}_k$ as \textit{spurious} or \textit{bias features} for $\mathcal{T}$ and refer to models' reliance on such features as \textit{prediction bias}.



%


\section{Background}
\label{sec:background}




\vspace*{-.3\baselineskip}
\paragraph{Spurious correlations of NLP datasets}
Previous work analyzing LLMs' error cases identified numerous false assumptions that LLMs use in prediction and can be misused to notoriously draw wrong predictions with the model.

In Natural Language Inference (NLI), where the task is to decide whether a pair of sentences entail one another, \citet{McCoy2019RightFT} identify LLMs' reliance on a lexical overlap and on specific shared syntactic units such as the constituents in the processed sentence pair.
\citet{nlp:asael-etal-2022-generative} identify the model's sensitivity to meaning-invariant structure permutations. 
Similarly, \citet{chaves-richter-2021-look} identify \textsc{BERT}'s reliance on the invariant morpho-syntactic composition of the input. 

In Question Answering, LLMs often rely on the positional relation of the question and possible answer words, such as assuming their close \textit{proximity} \citep{jia-liang-2017-adversarial}. \citet{Bartolo2020BeatTA} find that models tend to assume that questions and answers contain similar \textit{keywords}, remaining vulnerable to samples with none or multiple occurrences of the keywords in the context.
\citet{ko2020look} show models' preference for the answers in the first two sentences of the context, being statistically most likely to answer human-curated questions.

A perspective direction circumventing the biases introduced in data collection is presented in adversarial data collection \cite{jia-liang-2017-adversarial,Bartolo2020BeatTA} where the annotators collect the dataset with the intention of fooling the likely-biased model, possibly enhancing the model-in-the-loop in several fine-tuning iterations. Still, some doubts remain, as other work provides evidence that models trained on adversarial data may work better on adversarial datasets but underperform on other datasets \citep{kaushik-etal-2021-efficacy}, or introduce its own set of biases \cite{kovatchev-etal-2022-longhorns}. Nevertheless, our experiments (§\ref{sec:bias_ood_models}) show that training models on an adversarially-collected AdversarialQA dataset turns out to be among the most effective approaches to mitigating known prediction biases in question answering.

\paragraph{Debiasing methods}
A well-established line of work proposes to address the known dataset biases in the training process.
\citet{karimi-mahabadi-etal-2020-end} and \citet{he-etal-2019-unlearn} obtain a more robust, debiased model by (i)~training a \textit{biased model} that exploits the unwanted bias, followed by (ii)~training the debiased model as a complement to the biased one in a Product-of-Experts (PoE) framework \cite{Hinton2002TrainingPO}. \citet{clark2019don} extend this framework in the LearnedMixin method, learning to weigh the contribution of the biased and debiased model in the complementary ensemble. \citet{NEURIPS2021_878d5691} simulate the model for non-biased, out-of-distribution dataset through counterfactual reasoning \cite{DBLP:conf/cvpr/NiuTZL0W21} and use the resulting distribution for distilling target \cite{hinton2015distilling}, similarly to the LearnedMixin. 
Biased samples can also be identified in other ways, for instance, by the model's overconfidence \citep{wu-etal-2020-improving}.

In a complement to PoE approaches, other works apply model confidence regularization on the samples denoted as biased. \citet{feng-etal-2018-pathologies} and \citet{utama-etal-2020-mind} downweigh the predicted probability of the examples marked as biased by humans or a model. \citet{xiong2021uncertainty} find that a more precise calibration of the bias-detection model might bring further benefits to this framework, consistently with our observations (§\ref{sec:discussion_methods}).
Distributionally Robust Optimization (DRO) methods are another group of reweighting algorithms, addressing assumed imperfection of training datasets by (i)~segmenting data into \textit{groups} of diverse covariate shifts \citep{Sagawa*2020Distributionally} and (ii)~minimizing the worst-case risk over \textit{all} groups \citep{pmlr-v139-zhou21g}. We note that our bias measurement method closely relates to group DRO methods and can, for instance, serve as a method for quantifying per-group risk.

\vspace*{-.5\baselineskip}
\paragraph{Robustness measures}
Most of the work on enhancing models' robustness evaluates the acquired robustness on OOD datasets. 
In some cases, the evaluation utilizes datasets specially constructed to exploit the biases typical for a given task, such as HANS \cite{McCoy2019RightFT} for NLI, PAWS \cite{Zhang2019PAWSPA} for Paraphrase Identification, or AdversarialQA \cite{Bartolo2020BeatTA} for Question Answering, that we also use in evaluations.

Similar to us, some previous work quantified dataset biases by splitting data into two subsets, comparing model behaviour between these groups. \citet{McCoy2019RightFT} perform such evaluation over MNLI, demonstrating large margins in accuracy over the two groups and superior robustness of \textsc{BERT} over previous models.
Similarly, \citet{Utama2020TowardsDN} compare two groups based on prediction confidence.
Our Prediction bias measure follows a similar approach in QA but provides a more reliable assessment thanks to bootstrapping. Further, compared to the previous work, we assess models' reliance on a range of 7~spurious features, making our overall conclusions more robust.

An ability to measure a model's reliance on undesired features is also applicable in quantifying socially problematic biases. Previous work also utilizes specialized domain knowledge in models' bias evaluation but might not scale to other bias features; \citet{parrish-etal-2022-bbq} collect ambiguous contexts and assess the models' inclination to utilize stereotypes as prediction features.
\citet{bordia-bowman-2019-identifying} quantify LMs' gender bias by the co-occurrence of selected gender-associated words with gender-ambiguous words, such as \textit{doctor}.

\vspace*{-.3\baselineskip}
\section{Measuring Prediction Bias}
\label{sec:measusing}
\vspace*{-.3\baselineskip}

\begin{algorithm}[tb]
\SetKwProg{proc}{func}{}{}
\let\var\relax 
\centering
\scalebox{0.95}{%
\begin{minipage}{\linewidth}
\setstretch{1.1}
\proc{\upshape $\textit{measure\_bias}(\var{M}, \var{X}, \var{h}, \var{T_h})$:}{%
    $\var{A_h} \gets \var{h}(X)$ \\
    $\var{X_1} \gets \var{x_1} \in \var{X}: \var{A_h}(\var{x_1}) \leq \var{T_h}$ \\
    $\var{X_2} \gets \var{x_2} \in \var{X}: \var{A_h}(\var{x_2}) > \var{T_h}$ \\
    \ForEach{$\var{X_{1}'} \in \textit{repeat}(\textit{sample}(\var{X_1}))$}{
        $\var{E_1} \gets \var{E_1} + \textit{evaluate}(\var{M}\,(X_{1}'))$
    }
    \ForEach{$\var{X_{2}'} \in \textit{repeat}(\textit{sample}(\var{X_2}))$}{
        $\var{E_2} \gets \var{E_2} + \textit{evaluate}(\var{M}\,(X_{2}'))$
    }
    $\var{dist} \gets \max(0;\, E_1^\downarrow - E_2^\uparrow;\, E_2^\downarrow - E_1^{\uparrow})$ \\
    \Return{$\var{dist}$}
}
\vspace{1mm}
\end{minipage}
}
\caption{We measure \textit{Prediction bias} of the model $\var{M}$ exploited by the \textit{heuristic} $\var{h}$ on dataset $\var{X}$, as a \textit{difference} of $\var{M}$'s performance on two groups ($X_1$ and $X_2$) obtained by segmenting the samples of $X$ by the \textit{attribute} $\var{A_h} = \var{h}(X)$ on a given threshold~$\var{T_h}$.\hfill\newline
We bootstrap both evaluations, ($\textit{samples}=800$, $\textit{trials}=100$, and obtain two sets of measurements ($E_1$ and $E_2$), of which we subtract the upper and lower quantiles $E^\uparrow$ and $E^\downarrow$ ($q^\uparrow=0.975$, $q^\downarrow=0.025$) and consider such distance a scale of the learned prediction bias.
}
\label{algo:measure_bias}
\end{algorithm}

We assess a model's sensitivity to a known spurious feature in the following sequence of steps.
This methodology is visualized in Figure~\ref{fig:visual_abstract}, described in Algorithm~\ref{algo:measure_bias} and can be used to measure biases of any other QA model using the project repository\footnote{\url{https://github.com/MIR-MU/isbiased}}.

We start by (i)~implementing a \textit{heuristic}, i.e.\ a method $h: X \rightarrow \mathbb{R}$, that for \textit{all} samples of dataset~$X$ computes an \textit{attribute}~$A_h \in \mathbb{R}$ corresponding to the feature $\mathcal{F}$ that we suspise as non-representative, yet predictive for our training set and (ii)~we compute $h(x)$ for each sample $x$ of evaluation dataset~$X$. (iii)~We choose a threshold $T_h$ that we use to (iv)~split the dataset into two segments by $A_h$. Finally, (v)~we evaluate the assessed model $M$ on \textit{both} of these segments, in our case using Exact match evaluation, and (vi)~measure model \textbf{prediction bias} as the \textit{difference} in performance between these two groups. Using bootstrapped evaluation, we mitigate the effect of randomness by only comparing selected quantiles of confidence intervals. We propose to perform a hyperparameter search for the heuristic's threshold~$T_h$ that \textit{maximizes} the measured distance.


\vspace*{-.2\baselineskip}
\paragraph{Interpretation}
Given the reliance on bootstrapping, we state that the model's  \textit{true} performance polarisation is $0.975\times0.975 = 95.06\%$-likely to be equal or higher than the measured Prediction bias (with $q^\uparrow=0.975, q^\downarrow=0.025$ as in Algorithm~\ref{algo:measure_bias}).

Nevertheless, one should note that the proposed measure should not be used in a standalone but rather in a complement to an ID evaluation, as one can reduce the Prediction bias merely by \textit{lowering} the performance on the better-performing ID subset. Therefore, we report the values of Prediction bias together with the performance on a worse-performing, i.e. presumably non-biased split.

Another consideration concerns the ``natural'' polarisation of difficulty between samples; 
That is a portion of Prediction bias which can be explained by the features $\mathcal{F}$ that are \textit{representative} for the evaluated task (§\ref{sec:definitions}). 
One should note that the reduction of Prediction bias is meaningful only down to the level of the natural sample difficulty.

The validation set of SQuAD contains the annotations by three annotators that we use to quantify a level of Prediction bias that can be explained by the questions' natural difficulty (further denoted as \textit{Human} model); We report the minimum over Prediction biases of the annotators among each other.

Finally, even though we perform a hyperparameter search for optimal heuristics' thresholds $T_h$ feasible for a given size of dataset splits, there are no guarantees on the maximality of the found $T_h$.
Hence, Prediction bias only provides the \textit{lower bounds} of the model's polarisation.

\section{Experiments}
\label{sec:experiments}
Our main objective is to assess the efficiency of different training decisions in mitigating the reliance of the model on spurious correlations that can be present in datasets.
In Question answering, previous work identifies several spurious covariates in the SQuAD dataset \cite{rajpurkar-etal-2016-squad}; we build upon these findings and further extend the list of covariates learnable from SQuAD.

For each suspected bias feature, we first describe and implement the exploiting heuristics that we use to segment groups in the Prediction bias measure (§\ref{sec:biases}).
Subsequently, we observe the impact of the selected pre-training strategies (§\ref{sec:models}) and debiasing methods designed to address the over-reliance on biased features (§\ref{sec:mitigatingbias} -- §\ref{sec:assessedmethods}) on the Prediction bias and OOD performance of the resulting models. 

\subsection{Biases and Exploiting Heuristics}
\label{sec:biases}

Our work extends the list of previously reported QA biases based on our experience with two novel bias features that we later assess as significant.
The spurious features newly identified in this work are preceded with~\textbf{+}.

Together with each bias, we also briefly describe it's exploiting heuristic computing the non-representative feature $A_h$ (Algorithm~\ref{algo:measure_bias}).

\let\heuristic\paragraph
\heuristic{Distance of Question words from Answer words (\textit{word-dist})}
\label{subsec:worddistances}
\citet{jia-liang-2017-adversarial} propose that the models are prone to return answers close to the vocabulary of the question in context. 
Hence, \textit{word-dist} computes how close the closest question word is to the first answer in the context and computes the distance ($A_h$) as a number of words between the closest question word and the answer span.
\heuristic{Similar words between Question and Context (\textit{sim-word})}
\label{subsec:similarwords}
\citet{qa:shinoda-etal-2021-question} report the common occurrence of a high lexical overlap between the question and the correct answer over QA datasets.
In \textit{sim-word} heuristic, we represent the lexical overlap by the number of shared words between the question and the context. Both are defined as sets, and the intersection size of these two sets is computed as the heuristic's evaluation ($A_h$).

\heuristic{Answer position in Context (\textit{ans-pos})}
\label{subsec:kthsentence}
\citet{ko2020look} report that QA models may learn to falsely assume the answer's occurrence in the first two sentences. 
The exploiting heuristic first segments the context into sentences, and then identifies the sentence containing the answer and yields a scalar corresponding to the rank of the sentence within the context that contains the answer ($A_h$).

\heuristic{Cosine similarity of Question and Answer (\textit{cos-sim})}
\label{subsec:cosinesimilarity}
\citet{clark2019don} use the TF-IDF similarity as a biased model for QA, implicitly identifying a bias in undesired reliance of the model on the match of the keywords between the question and retrieved answer. 
We exploit this feature by (i)~fitting the TF-IDF model on all SQuAD contexts, (ii)~inferring the TF-IDF vectors of both questions and their corresponding answers, and (iii) returning the scalar ($A_h$) as cosine similarity between the TF-IDF vectors of question and answer.
\heuristic{Answer length (\textit{ans-len})}
\label{subsec:answerlength}
\citet{Bartolo2020BeatTA} show that QA models trained on SQuAD make errors much more often on questions asking for longer answers, implicitly identifying models' reliance on a feature that the answer must comprise at most a few words. 
We exploit this feature by simply computing $A_h$ as the length of the answer.
\heuristic{+Number of Question's Named Entities in Context (\textit{sim-ents})}
\label{subsec:similarentities}
We suspect that the in-context presence of multiple named entities, such as multiple personal names or locations, might perplex the QA model's prediction. This might suggest that models tend to reduce the QA task to a simpler yet irrelevant problem of Named Entity Recognition.
We utilize a pre-trained \textsc{BERT} NER model provided within \textsc{spaCy} library~\cite{spaCy} to identify named entities of the \textit{question type} (i.e., \textit{personal names} if the question starts with "Who"). Then, we count $A_h$ as the number of matching named entities in the context. 
\vspace*{-.4\baselineskip}
\heuristic{+Position of Question's subject to the~correct Answer in Context (\textit{subj-pos})}
\label{subsec:subjectposition}
Our observations suggest that the position of the question's subject in the context impacts the predicted answer spans of QA models.
In the corresponding heuristic, using \textsc{SpaCy} library, we (i)~identify the questions' subject expression and (ii)~locate its occurrences in the context. We (iii)~locate the answer span and compute $A_h$ as a relative position of the answer: either before the subject, after the subject, or after multiple occurrences of the question subject.

\subsection{Evaluated Models}
\label{sec:models}

To estimate the impact of selected pre-training strategies on the robustness of the resulting model, we conventionally fine-tune a set of diverse pre-trained LLMs for extractive~QA.

We alternate between the following models: \textsc{BERT-Base} \cite{devlin-etal-2019-bert}, \textsc{RoBERTa-Base} and \textsc{RoBERTa-Large} \cite{Liu2019RoBERTaAR}, \textsc{Electra-Base}\cite{Clark2020ELECTRAPT} and \textsc{T5-Large} \cite{t5}. This selection allows us to outline the impact of the various features on the robustness of the final QA model: (i) pre-training data volume (\textsc{BERT-Base} vs \textsc{RoBERTA-Base}), (ii) model size (\textsc{RoBERTA-Base} vs \textsc{RoBERTA-Large}), (iii) pre-training objective (\textsc{BERT-Base} vs \textsc{Electra-Base}), or (iv) extractive vs. generative prediction mode (\textsc{T5} vs. others).

We also evaluate the prediction bias of recent multi-task in-context learners, without fine-tuning: \textsc{T0} \cite{sanh2022multitask} trained for zero-shot in-context learning excluding SQuAD, 
and \textsc{Flan-T5} \cite{chung2022_flan} trained on a mixture of more than 1,800 tasks, including SQuAD.

\subsection{Debiasing Baseline: Resampling (\textsc{ReSam})}
\label{sec:mitigatingbias}

Based on the heuristics and their tuned configuration, our baseline method performs simple super-sampling of the underrepresented group ($X_1$ or $X_2$ in Algorithm~\ref{algo:measure_bias}) until the two groups are represented equally. This approach shows the possibility of bias reduction by simply normalizing the distribution of the biased samples in the dataset, requiring only the identification of the members of the under-represented group. \textsc{ReSam} closely follows the routine of Algorithm~\ref{algo:measure_bias} and splits the data by the optimal threshold of the attributes of the heuristics corresponding to each addressed bias.
\subsection{Assessed Debiasing Methods}
\label{sec:assessedmethods}

\begin{figure*}[tbh]
  \centerline{%
    \includegraphics[width=1.024\textwidth,
    keepaspectratio]{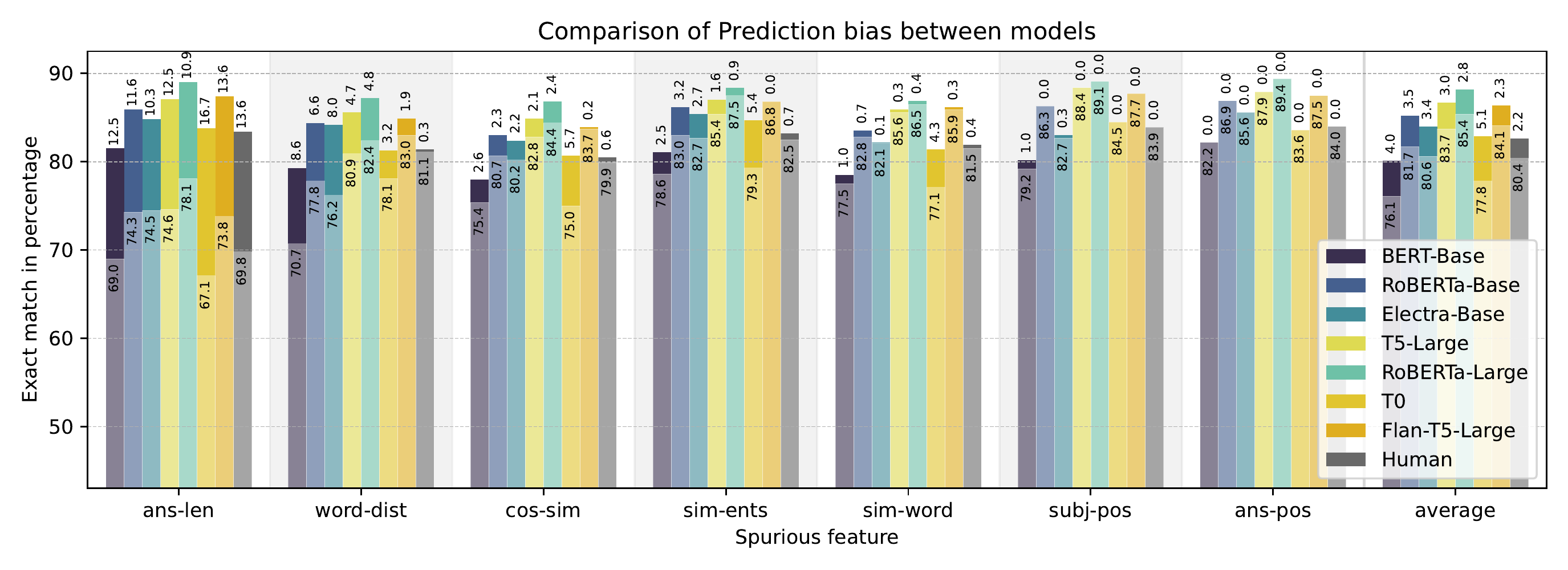}}
  \caption{\textbf{Prediction bias per pre-trained model.} The worse-performing split performance (lower bars) and Prediction bias (upper bars, sorted by group average) of QA models trained from different pre-trained LLMs, trained and evaluated on SQuAD for Exact match. Per-group bootstrapping of 100 repeats with 800 samples.
  }
  \label{fig:bias_between_models}
\end{figure*}

\begin{figure}[tbh]
  \centerline{%
    \includegraphics[width=1.05\textwidth,keepaspectratio]{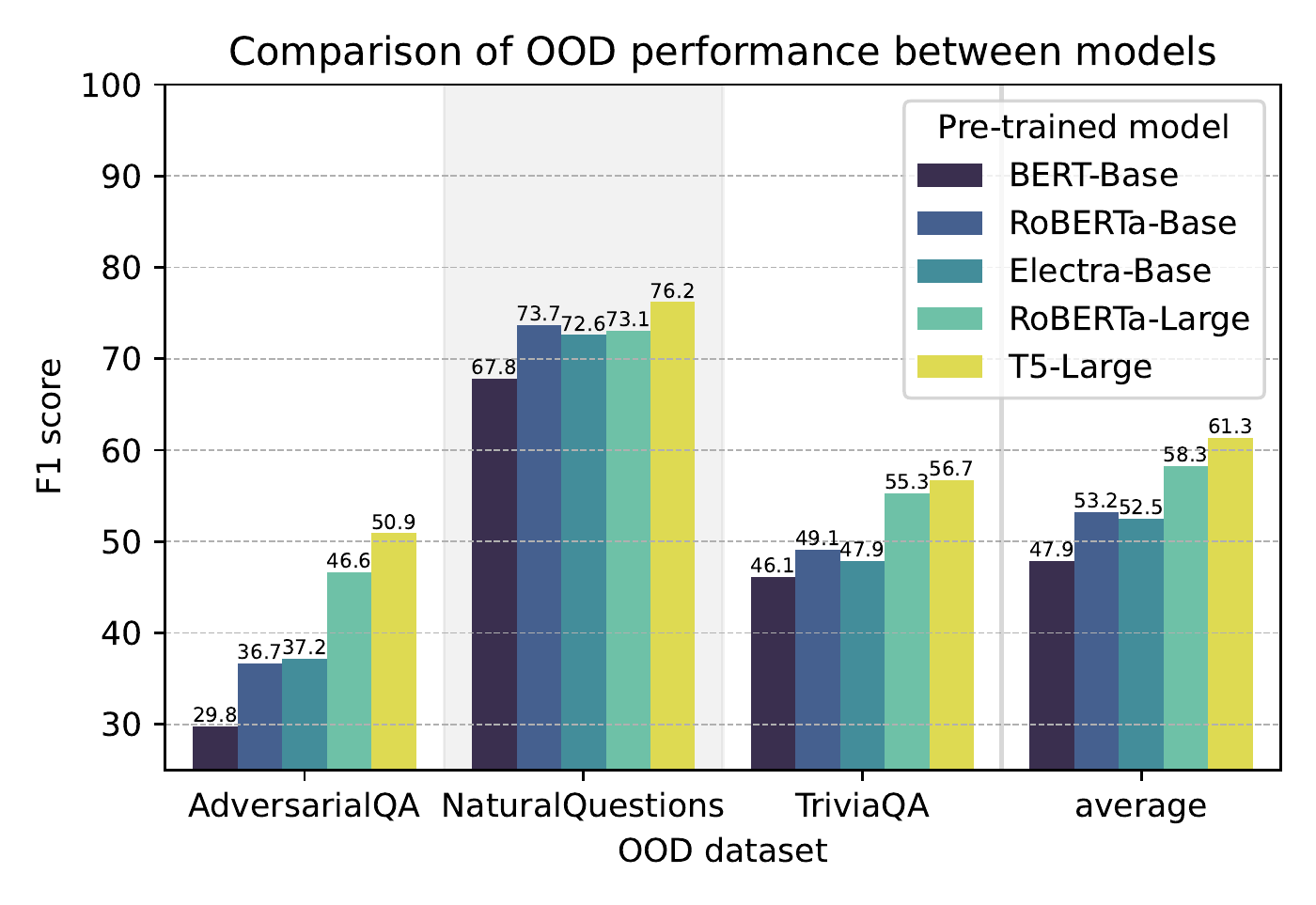}}
  \caption{\textbf{OOD performance per pre-trained model.} Comparison of F1-score of different models fine-tuned on SQuAD and evaluated on listed OOD datasets.
  }
  \label{fig:comparison_of_ood_performance}
\end{figure}

We assess the efficiency of debiasing methods in eliminating Prediction bias for the representatives of two diverse debiasing methods. In addition to Prediction bias, we also report the resulting performance on three OOD datasets. We follow the reference implementations as closely as possible while scaling the scope of experiments from one to seven separately-addressed biases. Complete description of training settings is in Appendix~\ref{apx:debiased_training}.

\paragraph{LearnedMixin (\textsc{LMix})} method \cite{clark-etal-2019-bert} is a popular adaptation of Product-of-Experts framework \cite{Hinton2002TrainingPO}, with a set of refinements (§\ref{sec:background}), that uses a \textit{biased model} as a complement of the trained debiased model in a weighted composition. We reimplement the reference implementation with the following alterations. Instead of the \textsc{BiDAF} model, we use stronger \textsc{BERT-Base} as the trained debiased model. Instead of using a TF-IDF-based bias model custom-tailored for a single bias type, we opt for a universal approach for obtaining biased models (Appendix~\ref{apx:bias_models}). We rerun the parameter search and choose a different \textit{entropy penalty} ($\textit{H}=0.4$) throughout all experiments.

\paragraph{Confidence Regularization (\textsc{CReg})} aims to reduce the model's confidence, i.e.\ the predicted score over samples marked as biased. \citet{utama-etal-2020-mind} propose to reduce the confidence of the biased samples using a distillation from the conventional QA teacher model, scaled down by the relative scores of a biased predictor. In our experiments, we consistently use \textsc{BERT-Base} for both the teacher and bias model. To enable comparability with \textsc{LMix}, we use identical bias models for both methods (Described in Appendix~\ref{apx:bias_models}). 


\begin{figure*}[tbh]
  \centerline{%
    \includegraphics[width=1.024\textwidth,keepaspectratio]{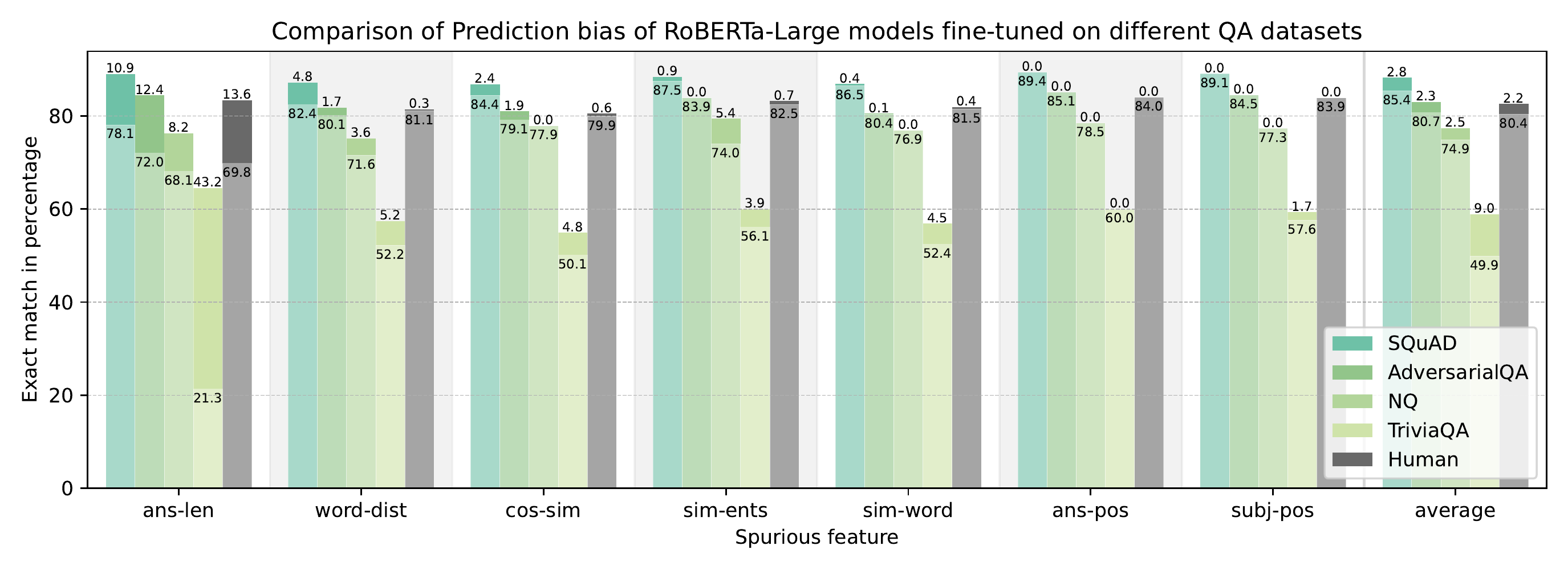}}
  \caption{\textbf{Prediction bias per dataset.} The worse-performing split performance (lower bars) and Prediction bias (upper bars) of \textsc{RoBERTa-Large} trained on different QA datasets, evaluated on a validation split of SQuAD for Exact match. All evaluation splits are identical, identified as maximal for the SQuAD-trained model (Appx.~\ref{apx:heuristics}).}
  \label{fig:bias_ood_models}
\end{figure*}

\section{Results}
\label{sec:results}

\subsection{Impact of Pre-training}
\label{sec:results_pretraining}

Figure \ref{fig:bias_between_models} compares the Prediction bias of the fine-tuned models of diverse pre-training data volumes and objectives, followed by in-context learning models and a human reference.

The results suggest that increased amounts of pre-training data of the base models (cf. \textsc{BERT-Base} and others) might mitigate the models' reliance on the bias. The results are less conclusive in a comparison of different pre-training objectives (cf. \textsc{RoBERTa-Base} and \textsc{Electra-Base}); While \textsc{Electra} is less polarised in 4 out of 7~cases, the differences are minimal. The largest reduction of Prediction bias ($-1.2$ on average) is achieved by increasing the model size of \textsc{RoBERTa-Large}.

Analogically, Figure~\ref{fig:comparison_of_ood_performance} compares OOD performance on selected QA datasets: AdversarialQA \cite{jia-liang-2017-adversarial}, NaturalQuestions \cite{kwiatkowski2019natural} and TriviaQA \cite{joshi2017triviaqa}. The concluding robustness ranking is mainly consistent with the Prediction bias ranking, with the exception of generative fine-tuning (\textsc{T5}), which outperforms others on OOD datasets but not on a reduction of the reliance on spurious features.

\subsection{Prediction bias of OOD models}
\label{sec:bias_ood_models}
Figure \ref{fig:bias_ood_models} compares Prediction bias over \textsc{RoBERTa-Large} models trained on different datasets. All evaluations are split on heuristics' thresholds $T_h$ optimal for the SQuAD model, which allows comparability to the shared human reference but implies that larger Prediction bias for OOD models might exist. We see that all Prediction biases learned on SQuAD are also learned from at least one OOD dataset. For the Trivia model, \textit{all} types of biases identified in SQuAD are magnified.

We specifically note the comparison of the Prediction bias of the SQuAD model to the model trained on AdversarialQA, collected adversarially to a SQuAD model. We find that the AdversarialQA model is the only OOD model lowering reliance on all biased features that are over the level of natural bias, supporting the argued efficiency of adversarial data collection in addressing original dataset biases.


\subsection{Impact of Debiasing}
\label{sec:impact_of_debiasing}

\begin{figure*}[tbh]
  \centerline{%
    \includegraphics[width=1.024\textwidth,keepaspectratio]{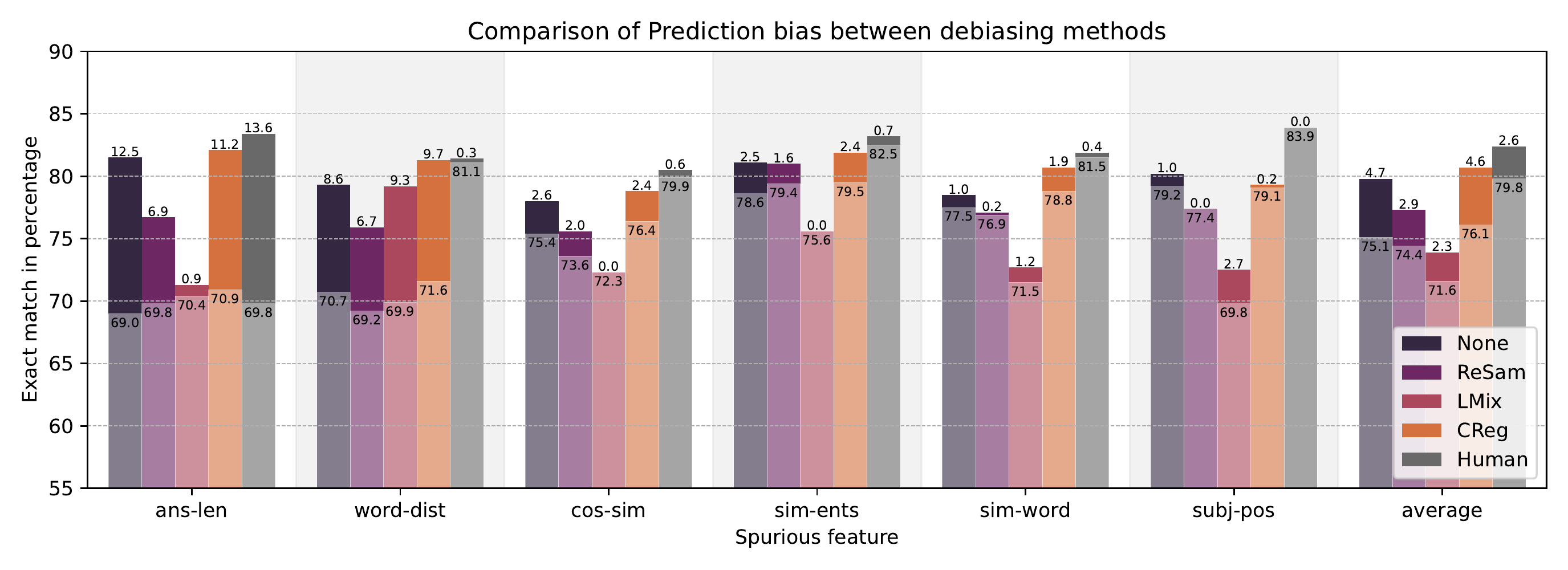}}
  \caption{\textbf{Prediction bias per debiasing methods.} The worse-performing split performance (lower bars) and Prediction bias (upper bars) of \textsc{BERT-Base} trained using selected debiasing methods, evaluated for Exact match on validation SQuAD. Per-group evaluations were measured using bootstrapping of 100 repeats with 800 samples.
  }
  \label{fig:bias_between_methods}
\end{figure*}

Figure \ref{fig:bias_between_methods} compares the biases of Question Answering models obtained within three debiasing methods (§\ref{sec:mitigatingbias} -- §\ref{sec:assessedmethods}), applied to the most-biased \textsc{BERT-Base} model.
We observe that debiasing methods are not consistent in the efficiency of mitigating the reliance on the addressed bias feature. In fact, only \textsc{ReSam} baseline lowers the bias of the original model consistently.
We attribute this inconsistency to methods' sensitivity to \textit{bias model}, further discussed in~§\ref{sec:discussion_methods}. While \textsc{LMix} is the most efficient in addressing Prediction bias in average, consistently to \citet{clark2019don} we see that this often comes for a price of the ID performance.

Table \ref{tab:ood_debiasing} enumerates the OOD performance of debiased models over three diverse QA datasets. By comparing these results to Prediction bias (Fig.~\ref{fig:bias_between_methods}), we see many cases where the reduction of Prediction bias can \textit{not} explain improvements of OOD; For instance, addressing \textit{word-dist} bias using \textsc{CReg} improves average F1-score on OOD datasets by $2.8$\% and by $7.5$ specifically on \textit{NaturalQuestions}, but the Prediction bias of such model increases by $1.1$ points. Similarly, \textsc{CReg} delivers $1.5$-point average gain of F1-score on OOD when addressing \textit{sim-word} bias but it also raises Prediction bias by $0.9$ points.

\begin{table}[t]
\maxsizebox{\linewidth}{\maxdimen}{
\begin{tabular}{@{}l@{\,}lll@{}}
\hline
 & \multicolumn{1}{l}{Original model} & \multicolumn{1}{c}{\!\!$29.8$ / $67.8$ / $46.1$} & \multicolumn{1}{c@{}}{} \\ \hline
 & \multicolumn{1}{c}{ReSam} & \multicolumn{1}{c}{LMix} & \multicolumn{1}{c}{CReg} \\
& \multicolumn{1}{c}{\!\!\textit{AQA} / \textit{\ NQ\ } / \textit{Trivia}} & \multicolumn{1}{c}{\!\!\textit{AQA} / \textit{\ NQ\ } / \textit{Trivia}} & \multicolumn{1}{c@{}}{\!\!\textit{AQA} / \textit{\ NQ\ } / \textit{Trivia}} \\ \hline

\textit{ans-len} & $-0.8$ / $-5.6 $ / $-1.7 $ & \!\!\!$-0.9$ / $-\!19.7$ / $-3.3$ & \!\!\!$-0.4$ / $+5.5 $ / $+\textbf{2.1}$\\
\textit{word-dist} & $+0.5$ / $+1.3$ / $+0.0$ & \!\!\!$+0.9 $ / $-\ 6.4 $ / $+1.5$ & \!\!$+\textbf{1.4}$ / $+\textbf{7.5} $ / $-0.5$\\
\textit{cos-sim} & $-0.1$ / $+0.3$ / $-1.3$ & \!\!\!$+0.4$ / $-\!11.3 $ / $-4.1$ & \!\!\!$-0.3$ / $+7.4 $ / $+1.1$\\
\textit{sim-ents} & $+1.1$ / $+1.5 $ / $+0.3 $ & \!\!\!$-0.1 $ / $\ -9.5 $ / $-1.2$ & \!\!\!$-1.0$ / $+5.9 $ / $+2.0$\\
\textit{sim-word} & $+0.3$ / $+0.1 $ / $+0.4$ & \!\!\!$-0.3 $ / $-\!21.4 $ / $-2.9$ & \!\!\!$-0.7$ / $+3.9 $ / $+1.4$\\
\textit{subj-pos} & $-1.6$ / $-0.7 $ / $-2.2 $ & \!\!\!$-1.3 $ / $-\!14.8 $ / $-1.3$ & \!\!\!$+0.0$ / $+5.1 $ / $+1.6$\\ \hline
\textit{Average} & \multicolumn{1}{c}{$-0.45$} & \multicolumn{1}{c}{\ $-5.31$} & \multicolumn{1}{c@{}}{\!$+2.33$}\\ \hline\vspace*{-10mm}
\end{tabular}%
}
\caption{\textbf{OOD performance of debiasing methods.} Differences of F1-scores of QA models trained on SQuAD using specified debiasing methods (§\ref{sec:assessedmethods}) to address selected bias features (§\ref{sec:biases}) evaluated on three OOD datasets; \textit{AdversarialQA} / \textit{NaturalQuestions} / \textit{TriviaQA}, respectively. Largest gains per dataset are in \textbf{bold}.}
\label{tab:ood_debiasing}
\end{table}

\begin{figure}[tbh]
  \centering
    \!\!\!\includegraphics[width=0.88\textwidth,keepaspectratio]{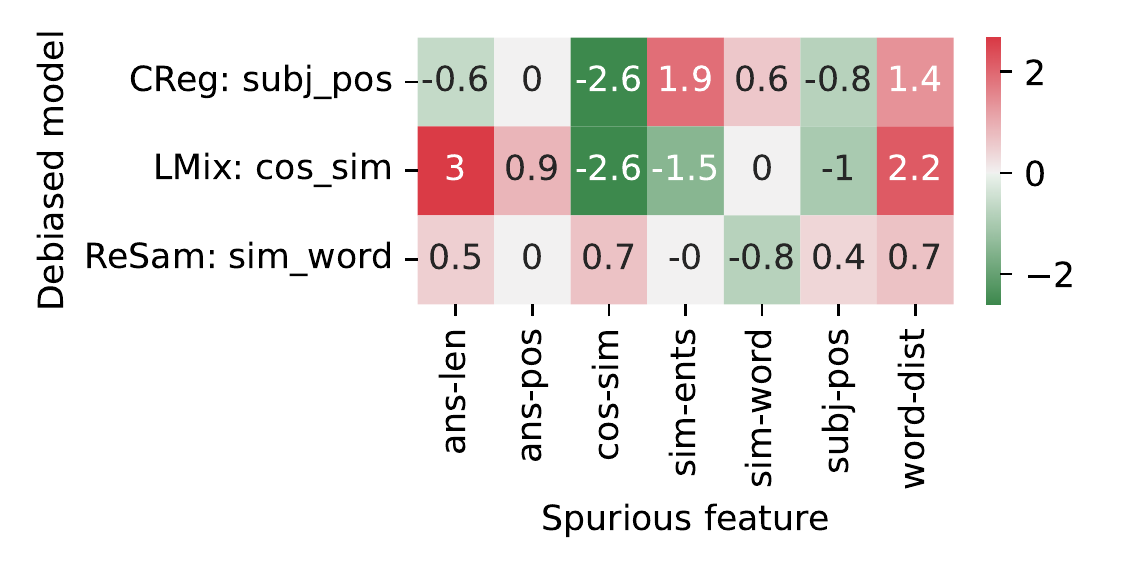}
  \caption{\textbf{Cross-bias evaluation of debiased models.} A relative change of Prediction bias by all spurious correlations, caused by applying inspected debiasing methods on \textsc{BERT-Base} QA model, in addressing specified spurious correlation. A full matrix is in Appx.~\ref{appx:cross_bias}, Fig.~\ref{fig:cross-bias_evaluation_all}.
  }
  \label{fig:cross-bias_evaluation}
\end{figure}

Figure \ref{fig:cross-bias_evaluation} further evaluates the impact of addressing one bias to other known biases in cases where each method delivers the largest Prediction bias reduction. 
We see that addressing a specific bias also affects the scope of the model's reliance on other covariates. 
Results suggest that \textsc{CReg} might be more robust to enlargening of other biases, increasing other Prediction biases by $0.31$ on average, as compared to \textsc{LMix} ($0.6$) and \textsc{ReSam} ($0.38$).

\looseness=-1
\section{Discussion}
\label{sec:discussion}


\paragraph{Pre-training and models' robustness}
The bias-level analyses of diverse pre-trained models (§\ref{sec:results_pretraining}) suggest that the mere increase of pre-training data and model parameters guide the fine-tuned models to lower reliance on biased features. However, we can find exceptions, such as in the case of \textsc{RoBERTa-Large} and \textsc{Electra-Base} on \textit{ans-len}. 
We speculate that even larger volumes of data might make the model more attracted to taking a shortcut through easier problem formulations, such as through Named entity recognition (cf. \textsc{BERT-Base} and \textsc{RoBERTa-Base} on \textit{sim-ents}).

Comparing the prediction bias of in-context learners with the fine-tuned models, we see that multi-task learning does not necessarily result in lower prediction bias or increased performance in the harder group; While \textsc{Flan-T5} on average reduces bias almost to the human level, \textsc{T0}'s quality is affected by spurious features even more than the models fine-tuned on biased SQuAD.

\paragraph{OOD performance and Prediction bias relation}
Our results conclude that the previously reported improvements in OOD performance attributed to the debiasing might not be attributed to the mitigated reliance on a spurious correlation; (i)~We measure that Prediction bias of the models trained directly on OOD datasets is still present over the level of human Prediction bias (§\ref{sec:bias_ood_models}). Therefore, it is possible to maintain OOD gains by learning to rely on biased features. (ii)~In practice, we find cases where applying a debiasing method \textit{magnifies} Prediction bias, but the resulting model still performs better in most OOD evaluations (§\ref{sec:impact_of_debiasing}).

\vspace*{-.2\baselineskip}
\paragraph{Practical aspects of applying debiasing methods}
\label{sec:discussion_methods}
\vspace*{-.2\baselineskip}

While we confirm that debiasing methods enable improvements in the OOD, we find that the significance of such improvements largely varies between the addressed biases, and the suitable configuration for one bias and dataset pair is often suboptimal for others. The scope of this variance can be seen in Table~\ref{tab:ood_debiasing} from the comparison of average OOD performance of \textsc{LMix} and \textsc{CReg} on \textit{word-dist}, used to pick methods' hyperparameters and bias models (Appendix \ref{apx:debiased_training}), and other biases; Both of the methods perform best on the bias used in parameter tuning, and the differences are often large. Bias-specific parameter tuning is further convoluted by the speed of the convergence of debiasing methods, which we measure as approximately 4~times slower for \textsc{CReg} and 3.5~times slower for \textsc{LMix}, compared to the standard fine-tuning of QA models.

The bias model is an important parameter of both assessed debiasing methods. We find that the scores have to be rescaled for trained bias models to avoid perplexing the trained model on biased samples and that the optimal scaling parameter is also bias-specific. 
The selection of the bias model also affects the optimal Entropy scaling $H$ of \textsc{LMix}; we find that the optimal value ($H=2.0$) for AdversarialQA reported by \textsc{LMix} authors is also not close to optimal ($H=0.4$) for our bias model.

\section{Conclusion}
\label{sec:conclusion}

Our work sets out to investigate the impact of various training decisions, including different pre-training and debiasing strategies, on models' reliance on specific spurious features in QA, complementing the commonly used out-of-distribution evaluations.
We use SQuAD to survey the existing and to identify new biased features but evaluate the reliance on these features for models trained on four different QA datasets.

We find that (i)~the OOD performance of different base models usually corresponds to models' reliance on bias features. However, (ii)~the state-of-the-art debiasing methods can improve OOD performance \textit{without} minimizing the model's reliance on spurious features, suggesting that dataset biases might be \textit{shared} among QA datasets. (iii)~We further evidence this by measuring the reliance on a spurious feature of models trained on other (OOD) datasets and find OOD models \textit{similarly} or even \textit{more reliant} on spurious features learnt from SQuAD.


We hope that our analyses will motivate future work to assess models' robustness also on a more detailed level of specific bias features, evading false conclusions on models' robustness, and, ultimately, accelerating progress towards creating more robust and reliable language models.

\section*{Limitations}

We highlight the limitation of our proposed evaluation method in the non-trivial \textit{interpretation} of the measured results, which we discuss in Section~\ref{sec:measusing}; We propose to measure the models' reliance on a bias feature as a difference of \textit{confidence intervals} of model performance on two data splits. This makes the conclusions about models' reliance (vs non-reliance) on a biased feature more robust, but it also perplexes the interpretation of measured absolute values. As a consequence, in the cases of different bias features ($\mathcal{F}_1$, $\mathcal{F}_2$) with very close prediction bias values, one should restrain from statements such as ``model M is more biased towards $\mathcal{F}_1$ than $\mathcal{F}_2$''.

We also underline that some biased features correlate with a \textit{natural} difference in the samples' difficulty. In such settings, a polarization of model performance might not be caused by its reliance on the spurious feature, but rather by other, natural features of the task. To disentangle the model's over-reliance on a biased feature from other aspects, we recommend contextualizing measured prediction bias with additionally measuring a \textit{human} level of prediction bias, that can be assessed on a set of duplicate annotations.

In our experiments, we measured considerable differences in natural difficulty only for a single feature – answer length – where it is likely more difficult to delimit the answer span for longer answers properly. We find that most models rely on this feature comparably to humans and refine our conclusions in Section~\ref{sec:results_pretraining} accordingly.

\section*{Acknowledgments}
We acknowledge the Centre for Biomedical Image Analysis at Masaryk University supported by MEYS CR (LM2023050 and CZ.02.1.01/0.0/0.0/18\_046/0016045 Czech-BioImaging) for their support in creating the models evaluated within this paper.


\bibliographystyle{acl_natbib}
\bibliography{lukasov,stefanik,sojkaglobal} 

\appendix 

\section{Cross-Bias Matrix of All Debiased Models}
\label{appx:cross_bias}

\begin{figure}[tb]
  \centering
    \!\!\!\!\includegraphics[width=1.056\textwidth,keepaspectratio]{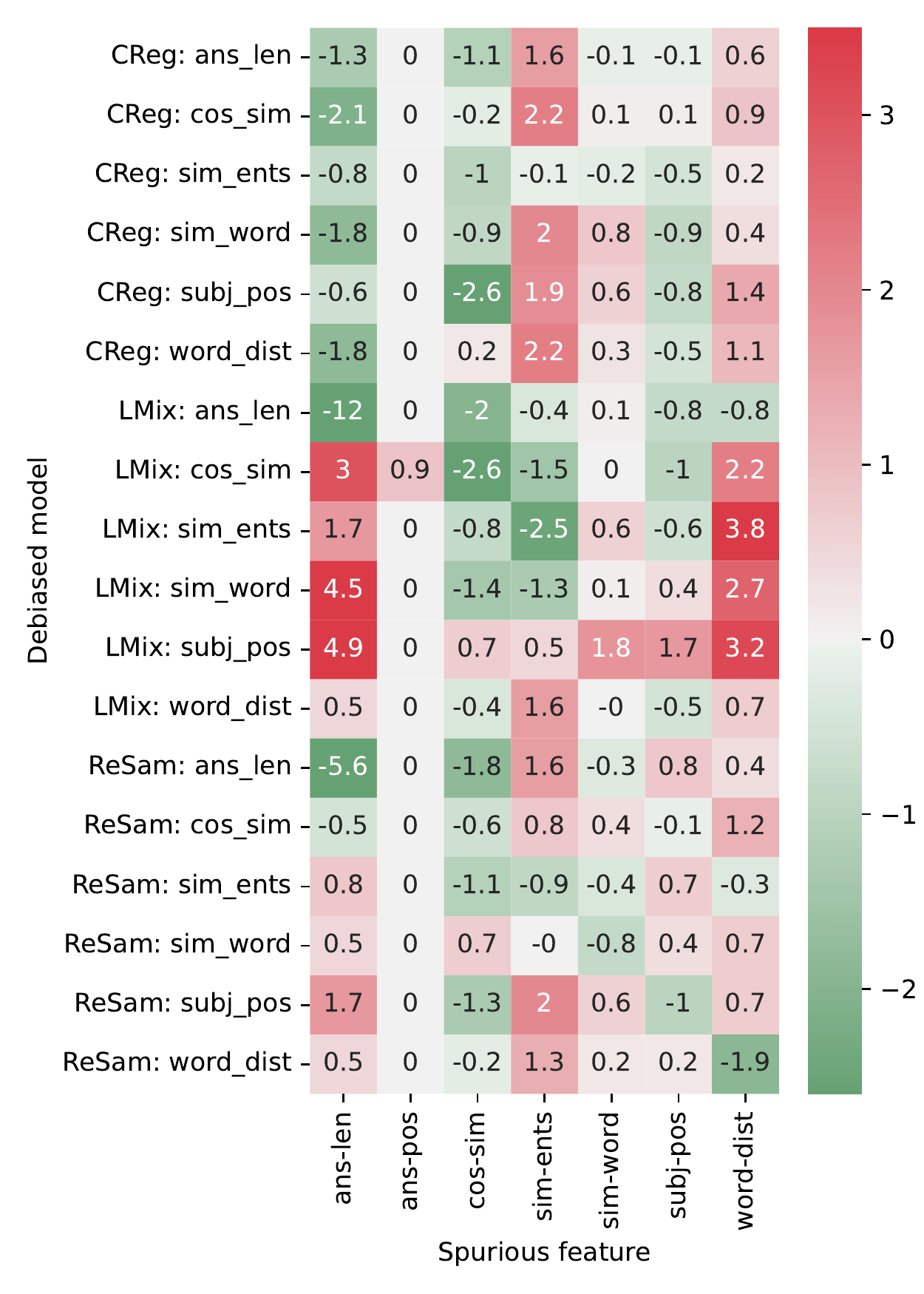}
    \vspace*{-5mm}
  \caption{\textbf{Full cross-bias evaluation of debiased models.} A relative change of Prediction bias by all spurious correlations, caused by applying inspected debiasing methods on \textsc{BERT-Base} QA model, in addressing specified spurious correlation.}
  \label{fig:cross-bias_evaluation_all}
\end{figure}

Figure \ref{fig:cross-bias_evaluation_all} shows the change of Prediction bias by applying the listed debiasing methods to eliminate the associated bias feature. We see that some biases are more difficult to address, while other ones can be transitively addressed through others.

\section{Details of Training Configurations}
\label{apx:training_config}

This section overviews all configurations that we have set in training the debiased models (§\ref{sec:mitigatingbias} -- \ref{sec:assessedmethods}) as well as the conventional QA fine-tuning comparing the impact of pre-training on QA models' robustness (§\ref{sec:models}).

\subsection{Standard Fine-tuning}
\label{apx:conventional_training}

For model fine-tuning, we use following hyperparameters: \textbf{learning rate:} 2e$^{-5}$, \textbf{batch size:}~16, \textbf{evaluation:} each 200~steps and \textbf{train epochs:}~3. We also set the \textbf{early stopping patience} to 10~evaluation steps, based on a validation loss of the training dataset (SQuAD) also used for selecting the evaluated model. The \textbf{validation loss} of the evaluated model is 1.02. All other parameters can be retrieved from the defaults of TrainingArguments of HuggingFace~\cite{Wolf2019HuggingFacesTS} in version 4.19.1.

We use the listed configuration also in training the generative \textsc{T5} model. We use the Adaptor library \cite{stefanik-etal-2022-adaptor} in version 0.1.6 for fine-tuning \textsc{T5} for generating answers.

\subsection{Debiasing Training Experiments}
\label{apx:debiased_training}

\subsubsection{Bias models}
\label{apx:bias_models}

The canonical debiasing implementations utilize bias-specific models for identifying bias; \citet{clark-etal-2019-bert} use the TF-IDF model as a scalar of possible bias for each QA sample, while \citet{utama-etal-2020-mind} experiment with a percentage of the shared words and cosine embeddings between word distances, in NLI context. 

As we scale our experiments to six different biases, we opt for a universal approach for obtaining bias models for both \textsc{LMix} and \textsc{CReg} and train each biased model on a better-performing segment of the dataset identified using the approach described in Section~\ref{sec:measusing}.
For all our biased models, we train \textsc{BERT-Base} architecture from scratch and pick the checkpoint with a maximal difference of the F1-score between the two segments from the validation split of SQuAD. 


While our approach scales well over many biases, a significant difference between the learned bias models original ones, such as TF-IDF, is the \textit{scale} of prediction probabilities; As the trained bias models become very confident on a biased subset, often reaching probabilities close to 1 for the biased samples. A ``perfect'' bias model causes problems for both \textsc{LMix} and \textsc{CReg} as such model forces the trained model to avoid correct predictions on the biased samples completely. We learn to address this problem by rescaling bias predictions and tuning the scaling interval based on a validation performance of the debiased model. 
Consequently, we scale the bias probabilities to $\langle0; 0.2\rangle$ for \textsc{LMix} and $\langle0; 0.1\rangle$ for \textsc{CReg}.
Further details on bias models can be found in Appendix~\ref{apx:debiased_training}.

In the initial phase, we experiment with diverse configurations and sizes of bias models, intending to maximize the polarization of performance on the biased and non-biased subsets. Among different configurations of model sizes and configurations, we find that the highest polarisation can be reached using \textsc{BERT-Base} architecture trained from scratch. We fix this decision and the parameters (learning rate 4e$^{-5}$, a number of training steps 88,000) with respect to the maximum OOD (AdversarialQA) F-score of this model of \textsc{LMix} 
model addressing \textit{word-dist} bias. Our bias models reach between 18\% and 59\% of accuracy on easier, i.e., biased data split while between 4\% and 19\% on the non-biased one.

\subsubsection{Baseline debiasing: Resampling}
We train the \textsc{ReSam} analogically to Baseline Fine-tuning experiments (§\ref{apx:conventional_training}).
Compared to other debiasing methods, \textsc{ReSam} baseline is non-parametric, including no dependence on the bias model. 

Even though we find \textsc{ReSam} to be the only method mitigating Prediction bias in all the cases, our further analyses show that its enhancements on OOD datasets vary among biases. Figure~\ref{fig:loss_base_distances} shows validation losses from the training on SQuAD resampled using \textsc{ReSam} by \textit{word-dist}, while analogically, Figure~\ref{fig:loss_base_entities} shows the losses for \textit{sim-ents} bias. While in the former case, \textsc{ReSam} does not stably reach lower loss on OOD datasets, in the latter case, validation losses are consistently lower between steps 7,000 and 8,000, where the SQuAD validation loss used to pick the best-performing model plateaus.

\subsubsection{Learned Mixin}

In addition to the implementation and default parameters of \citet{clark2019don}, we find that the additional entropy regularization component \textit{H} makes a significant difference in the resulting model evaluation. Therefore we perform a hyperparameter search over the values of \textit{H} used for QA by \citet{clark2019don} on \textit{word-dist} bias, optimizing the OOD performance on AdversarialQA \cite{Bartolo2020BeatTA} and eventually fix $\textit{H}=0.4$ over all our experiments. 

Following the low initial OOD performance of \textsc{LMix} as compared to the results of \citet{clark2019don}, we further investigate covariates of this result and identify \textsc{LMix}'s high sensitivity to bias model; while in the original implementation, TF-IDF similarities of question and answer segment likely never reach $1.0$, our generic bias models reaches $1.0$ probability for most of the samples marked as biased. Hence, we introduce a parameter of scaling interval $\langle0; x\rangle$ of bias model's scores, where we optimize $x \in \langle0.2; 0.4; 0.5; 0.6; 0.7; 0.8; 0.9; 0.95\rangle$ according to the maximum ID F-score of the debiased model addressing \textit{word-dist} bias, fixing optimal $x=0.8$ throughout all other experiments. All other parameters remain identical to the standard fine-tuning (§\ref{apx:conventional_training}).


\subsubsection{Confidence Regularization}

While the authors of \textsc{CReg} \cite{utama-etal-2020-mind} find benefits in its non-parametricity, we find that \textsc{CReg} also shows high sensitivity to a selection of bias model, guiding us to also rescale the prediction of the bias model in the training distillation process. We use the same methodology to pick the scaling interval $\langle0; x\rangle$ for \textsc{CReg} as for \textsc{LMix} and fix $x=0.9$ as the optimal one. All other parameters remain the identical to the standard fine-tuning (§\ref{apx:conventional_training}). 

We implement \textsc{CReg} using Transformers library \cite{wolf2020transformers} in version 4.19.1.

\section{Exploiting Heuristics Configuration}
\label{apx:heuristics}

Here we enumerate the optimal thresholds over all pairs of the implemented heuristics, as picked according to \textsc{BERT-Base-Cased} model.

We assess the candidate thresholds among all possible values within the range of the computed values $A_h$ computed over $X=\text{SQuAD}_\text{valid}$ (see Algorithm~\ref{algo:measure_bias}), with steps of 1 for possible values higher than 1 and 0.1 for values between 0 and 1, within the valid interval; We set the validity interval such that the resulting splits of the dataset must each have a size of at least two times of the sample size parameter, except where there is only one significant threshold, and its size is larger than the sample size. The optimal threshold value is then the one that delivers the highest Prediction bias value. We find and use the following optimal thresholds of \textsc{BERT-Base-Cased} evaluated on $X=\text{SQuAD}_\text{valid}$ for specific biases: 7~for \textit{word-dist}, 3~for \textit{sim-word}, 4~for \textit{ans-len}, 0.1 for \textit{cos-sim}, 0~for \textit{sim-ents} and 1~for \textit{subj-pos}. A corresponding number of samples in the underperforming groups of $\text{SQuAD}_\text{valid}$ (n=10,570) are following: 1,651 for \textit{word-dist}, 3,281 for \textit{sim-word}, 3,124 for \textit{ans-len}, 954 for \textit{cos-sim}, 5,006 for \textit{sim-ents} and 1,672 for \textit{subj-pos}.

The implementations of some biases' heuristics utilize external libraries for entity recognition or TF-IDF vectorization. For these, we used \textsc{SpaCy} in version 3.4.1 and \textsc{NLTK} in version 3.4.1.

\section{Experimental Environment}
\begin{figure*}[tbh]
  \centering
    \includegraphics[width=\textwidth,keepaspectratio]{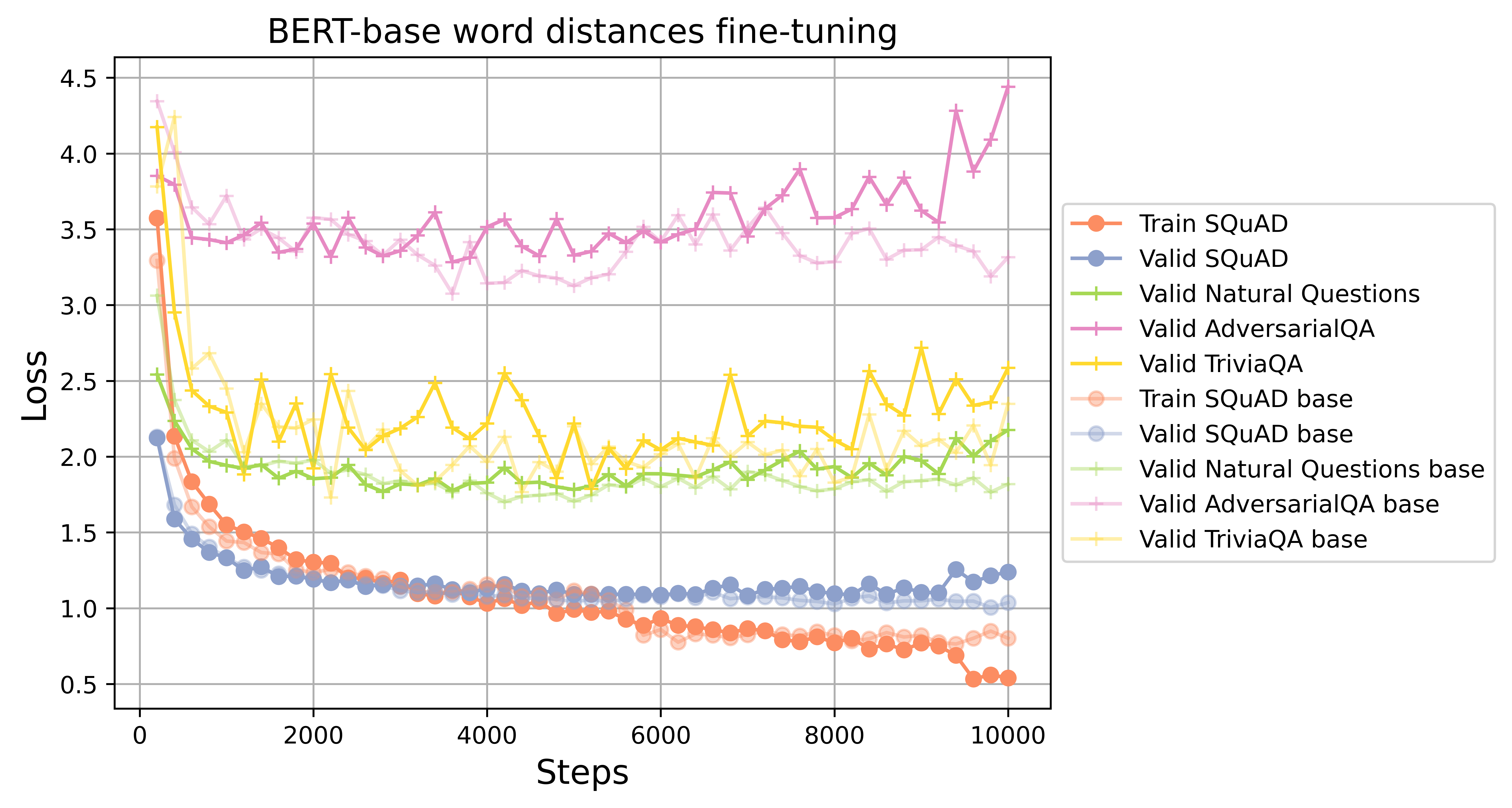}
  \vspace*{-\baselineskip}
  \caption{Development of validation loss of \textbf{\textsc{ReSam}} addressing \textbf{\textit{word-dist}} bias (darker plots) and standard fine-tuning (lighter plots) for Question Answering on SQuAD, also evaluated on other (OOD) datasets, for the first 10,000 steps.}
  \label{fig:loss_base_distances}
\end{figure*}

\begin{figure*}[tbh]
  \centering
    \includegraphics[width=\textwidth,keepaspectratio]{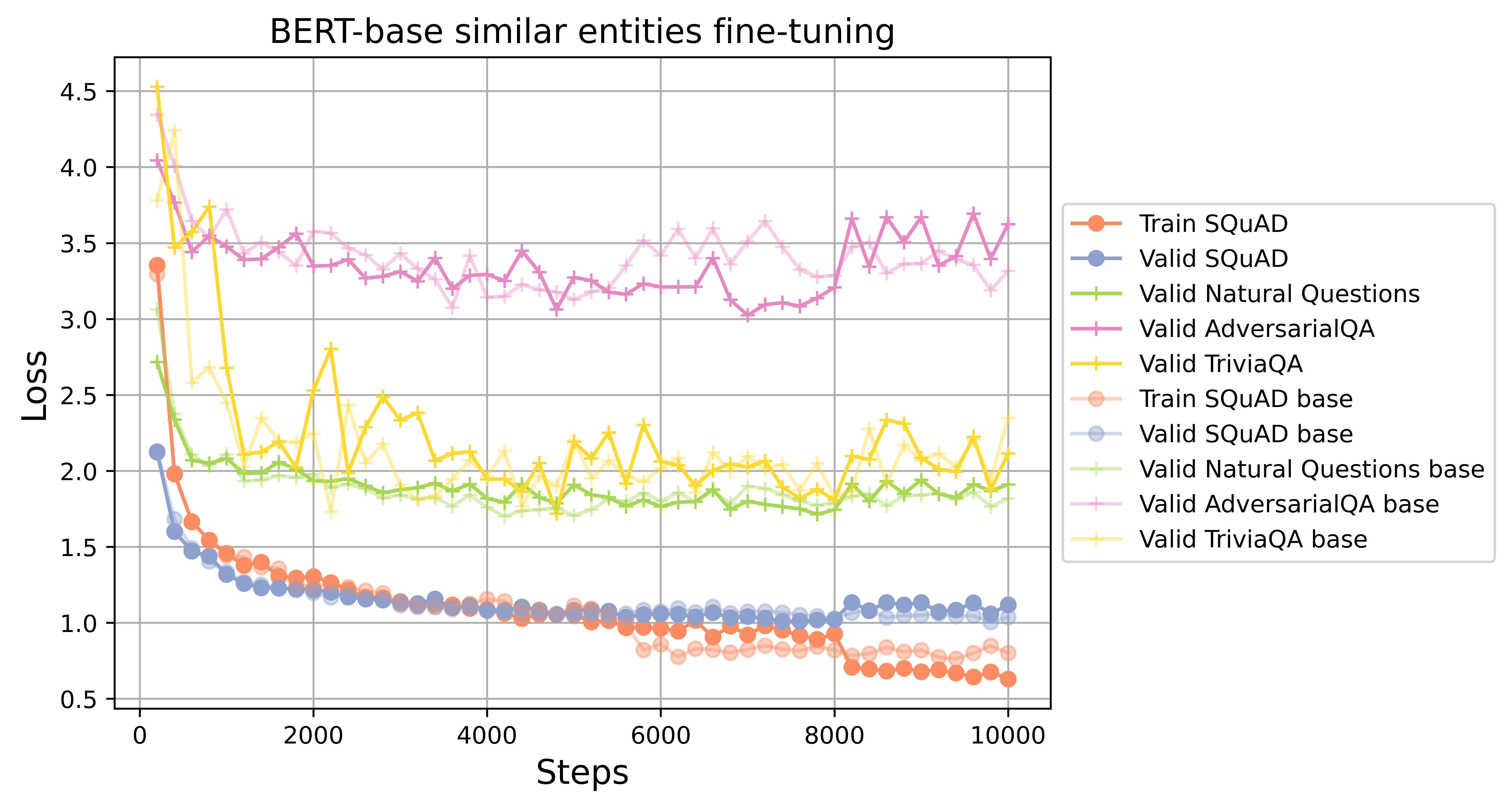}
  \vspace*{-\baselineskip}
  \caption{Development of validation loss of \textbf{\textsc{ReSam}} addressing \textbf{\textit{sim-ents}} bias (darker plots) and standard fine-tuning (lighter plots) for Question Answering on SQuAD, also evaluated on other (OOD) datasets, for the first 10,000 steps.}
  \label{fig:loss_base_entities}
\end{figure*}

Our experiments utilized a single NVidia A100 GPU with 80\,GB of VRAM, a single CPU core, and less than 32\,GB of RAM. However, all our experiments can be run using a lower compute configuration, given a longer compute time; The inference of a single-sample prediction batch of \textsc{RoBERTa-Large} as our largest model requires only 13\,GB of VRAM.
The debiasing training runs take longer to converge, as compared to standard fine-tuning; While the conventional training and \textsc{ReSam} converge within 10,000 steps (Figures~\ref{fig:loss_base_distances} and \ref{fig:loss_base_entities}) we find that \textsc{LMix} requires between 60,000 and 100,000 steps, and \textsc{CReg} needs between 20,000 and 30,000 steps to converge, making the debiasing training 4--8~times slower in average. In our training configuration, each of the reported training runs takes between 50~minutes and 1~hour per 10,000 updates. Given that our evaluation already aggregates the bootstrapped results, we perform a single run for each experiment, which might result in a wider confidence interval and consistently smaller measured volumes of Prediction bias.




\end{document}